\documentclass[10pt,twocolumn,letterpaper]{article}

\usepackage{iccv}
\usepackage{times}
\usepackage{epsfig}
\usepackage{graphicx}
\usepackage{amsmath}
\usepackage{amssymb}
\usepackage{mathtools}
\usepackage{bbm}
\usepackage{bm}
\usepackage{hhline}
\usepackage{ctable}
\usepackage{algorithm}
\usepackage{algpseudocode}
\usepackage[noadjust]{cite}
\usepackage{enumitem}

\usepackage[pagebackref=true,breaklinks=true,letterpaper=true,colorlinks,urlcolor=black,bookmarks=false]{hyperref}
\hypersetup{citecolor=[RGB]{119,185,0}}

\iccvfinalcopy 

\def\iccvPaperID{3715} 

\ificcvfinal\fi

\begin{document}

\title{\textsc{DiscoBox}: Weakly Supervised Instance Segmentation and Semantic Correspondence from Box Supervision}

\author{
Shiyi Lan$^{1}$\thanks{Work done during an internship at NVIDIA Research.}
~~~\,Zhiding Yu$^{2}$\thanks{Corresponding author: \textless{}\href{mailto:zhidingy@nvidia.com}{zhidingy@nvidia.com}\textgreater{}.}
~~~\,Christopher Choy$^{2}$
~~\,Subhashree Radhakrishnan$^{2}$
~~\,Guilin Liu$^{2}$\\
~~\,Yuke Zhu$^{2,3}$
~~\,Larry S. Davis$^{1}$
~~\,Anima Anandkumar$^{2,4}$\\
$^1$University of Maryland, College Park
~~\,$^2$NVIDIA
~~\,$^3$The University of Texas at Austin
~~\,$^4$Caltech
}

\maketitle
\ificcvfinal\thispagestyle{empty}\fi

\begin{abstract}
We introduce DiscoBox, a novel framework that jointly learns instance segmentation and semantic correspondence using bounding box supervision. Specifically, we propose a self-ensembling framework where instance segmentation and semantic correspondence are jointly guided by a structured teacher in addition to the bounding box supervision. The teacher is a structured energy model incorporating a pairwise potential and a cross-image potential to model the pairwise pixel relationships both within and across the boxes. Minimizing the teacher energy simultaneously yields refined object masks and dense correspondences between intra-class objects, which are taken as pseudo-labels to supervise the task network and provide positive/negative correspondence pairs for dense constrastive learning. We show a symbiotic relationship where the two tasks mutually benefit from each other. Our best model achieves 37.9\% AP on COCO instance segmentation, surpassing prior weakly supervised methods and is competitive to supervised methods. We also obtain state of the art weakly supervised results on PASCAL VOC12 and PF-PASCAL with real-time inference.
\vspace{-0.5cm}
\end{abstract}

\section{Introduction}

The ability to localize and recognize objects is at the core of human vision. This has motivated the vision community to study object detection~\cite{viola2001rapid} as a fundamental visual recognition task. Instance segmentation~\cite{hariharan2014simultaneous} is further introduced on top of detection to predict the foreground object masks, thus enabling localization with pixel-level accuracy. More recently, a growing number of works aim to lift the above tasks to the 3D space~\cite{hejrati2012analyzing,xiang2014beyond,xiang2016objectnet3d,kundu20183d,gkioxari2019mesh}. As a result, landmark~\cite{hejrati2012analyzing,bulat2017far} and (semantic) correspondence~\cite{lowe2004distinctive,bay2006surf,tola2009daisy,rosten2006machine,verdie2015tilde,han2015matchnet,yi2016lift,ono2018lf,sarlin2020superglue,liu2010sift,zhou2015flowweb,choy2016universal,rocco2017convolutional,min2019hyperpixel,liu2020semantic,kulkarni2019canonical,rocco2018end,chen2019show,hung2019scops,lee2019sfnet,jeon2020guided,min2020learning} have been widely studied to associate object parts across different views. These methods have become critical components in pose estimation~\cite{tulsiani2015viewpoints,collet2011moped,pavlakos20176,tremblay2018deep} and reconstruction~\cite{snavely2006photo,kanazawa2018learning,novotny2019c3dpo,li2020self} because they help to reduce uncertainties through additional constraints, such as determining camera poses and viewpoints~\cite{kanazawa2018learning,tulsiani2015viewpoints}.

\begin{figure}[t]
\centering
\includegraphics[width=1.0\linewidth]{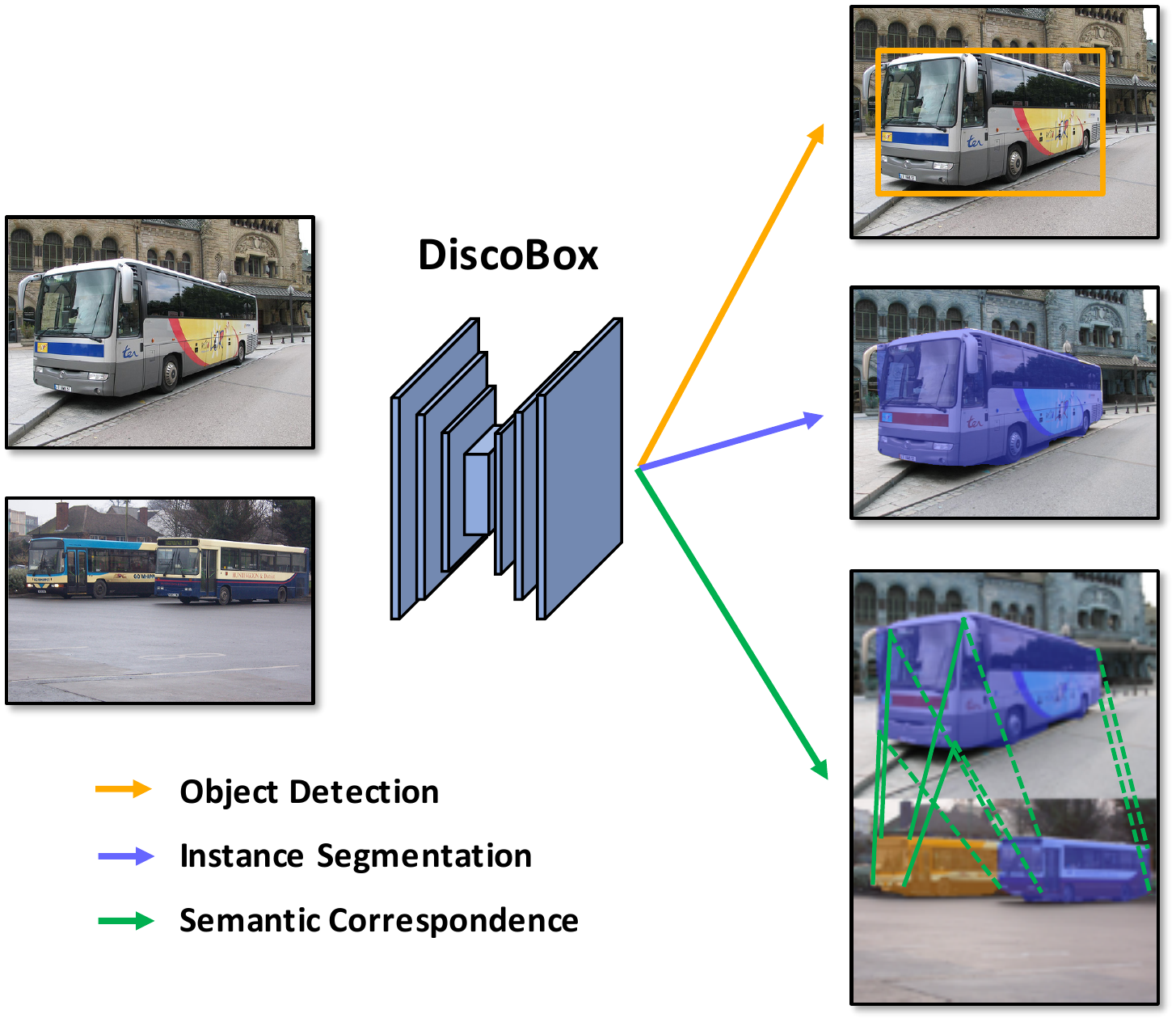}
\caption{Task overview. Given a pair of images, \textsc{DiscoBox} simultaneously outputs detection, instance segmentation and semantic correspondence predictions. Best viewed in color.}
\label{fig:teaser}
\end{figure}

Among various correspondence tasks, semantic correspondence aims to establish the associations across different scenes and object instances, and is arguably the most challenging one due to large variations in appearance and pose. The literature of semantic correspondence and instance segmentation have largely remained decoupled. For instance, the main semantic correspondence benchmarks~\cite{WahCUB_200_2011,zhou2015flowweb,taniai2016joint,novotny2016have,ham2017proposal,min2019spair} have been focusing on object-centric scenarios which de-emphasizes the role of object localization, while the latest instance segmentation methods do not make use of intra-class correspondences. However, these seemingly separate problems can benefit from each other because associating object parts requires understanding the object of interest a priori. Similarly, knowing the semantic parts of an object requires understanding the geometry of functional parts and can improve object localization~\cite{zhu2012face,huang2015densebox}.

Even though the advantage of learning correspondences and instance segmentation jointly is clear, many state of the art methods do not make use of this approach due to the lack of large scale datasets with both masks and correspondences. To overcome this challenge, weakly supervised methods have been recently introduced to relax the need for costly supervision in both tasks~\cite{khoreva2017simple,hsu2019weakly,arun2020weakly,tian2021boxinst,hung2019scops,rocco2018end,chen2019show,lee2019sfnet,jeon2020guided,min2020learning}. Our work is aligned with these efforts as we aim to address instance segmentation and semantic correspondence jointly with inexpensive bounding box supervision. This allows us to effectively push the boundaries with more data.

More importantly, box supervision presents a principled way to couple the above two tasks: First, instance segmentation greatly extends the capability of semantic correspondence to handle multi-object scenarios. This allows one to define a more generalized and challenging semantic correspondence task where the performance emphasizes both the quality of object-level correspondence and the accuracy of object localization. Second, multi-tasking provides the mutual constraints to overcome trivial solutions in box supervision. Indeed, our study shows a symbiotic relation where localization benefits correspondence via improved locality and representation, whereas correspondence in turn helps localization with additional cross-image information.

We propose \textsc{DiscoBox}, a framework which instantiates the above targets as shown in Fig.~\ref{fig:teaser}. \textsc{DiscoBox} leverages various levels of structured knowledge and self-supervision both within and across images to reduce the uncertainties.

\vspace{0.15cm}
\noindent\textbf{Summary of contributions:}
\begin{itemize}[leftmargin=*]
    \item Our work is the first to propose a unified framework for joint weakly supervised instance segmentation and semantic correspondence using bounding box supervision.
    \item We propose a novel self-ensembling framework where a teacher is designed to promote structured inductive bias and establish correspondences across objects. We show that the proposed framework allows us to jointly exploit both intra- and cross-image self-supervisions and leads to significantly improved task performance.
    \item We achieve state-of-the-art performance on weakly supervised instance segmentation. Our best model achieves 37.9\% AP on COCO test-dev, surpassing competitive supervised methods such as YOLACT++~\cite{bolya2020yolact++} (34.6\% AP) and Mask R-CNN~\cite{xie2020polarmask} (37.1\% AP).
    \item We also achieve state-of-the-art performance on weakly supervised semantic correspondence, and are the first to propose a multi-object benchmark for this task.
\end{itemize}

\textbf{Task network.} Our task network contains an instance segmentation backbone with a multiple instance learning head. The module is supervised by bounding boxes which contain rich object information. Through multiple instance learning, coarse object masks naturally emerge as network attention, and is taken by the teacher as initial predictions.

\textbf{Teacher model.} The teacher is defined by a Gibbs energy which comprises a unary potential, a pairwise potential and a cross-image potential. The unary potential takes the initial output from the student whereas the pairwise and cross-image potentials model the pairwise pixel relationships both within and across bounding boxes. Minimizing the teacher energy promotes contrast-sensitive smoothness while establishing dense correspondence across the objects. This allows one to consider cross-image self-supervision where correspondence provides positive and negative pairs for dense contrastive learning. We show that this in turn can improve the quality of instance segmentation.

Our promising results indicate the possibility to completely remove mask labels in future instance segmentation problems. We also envisage the wide benefit of \textsc{DiscoBox} to many downstream applications, particularly 3D tasks.
\section{Related Work}

\subsection{Object recognition and localization}

\textbf{Object detection.} Object detection has been an active research area with rich literature. Training on large amounts of bounding box annotations with convolutional neural networks (CNNs) has become a standard paradigm~\cite{girshick2014rich}. Initial CNN based detectors tend to share a multi-stage design~\cite{girshick2014rich,ren2015faster} where the first stage gives redundant object proposals, followed by refinement by CNNs in the second stage. A recent trend of design aims to reduce the complexity by having one-stage architectures~\cite{redmon2016you,liu2016ssd,zhou2019objects,tian2019fcos}, and therefore achieves good trade-off between efficiency and performance. Our weakly supervised design allows~\textsc{DiscoBox} to be conveniently trained like any object detection algorithm on the increasingly large datasets~\cite{everingham2010pascal,lin2014microsoft,shao2019objects365}, but output additional predictions beyond just bounding boxes.

\textbf{Instance segmentation.} Instance segmentation aims to produce more precise localization over detection by predicting the object segmentation masks. Bharath et al.~\cite{hariharan2014simultaneous} are the first to introduce an R-CNN-based framework with a precision-recall benchmark. Similar to R-CNN~\cite{girshick2014rich}, their object proposal and mask generation~\cite{arbelaez2014multiscale} is not end-to-end learnable. Recent methods including Mask R-CNN~\cite{dai2016instance,li2017fully,he2017mask} have largely followed this ``detection-flavored'' design and benchmarking, but introduce end-to-end learnable object proposal and mask prediction. Inspired by the one-stage detection, a number of one-stage instance segmentation methods have also been proposed~\cite{bai2017deep,bolya2019yolact,bolya2020yolact++,tian2020conditional,xie2020polarmask,wang2020solov2}. These methods all require mask annotations during training, whereas \textsc{DiscoBox} only needs box labels. \textsc{DiscoBox} is also agnostic to the choice of frameworks. In this work, we showcase \textsc{DiscoBox} on both YOLACT++~\cite{bolya2020yolact++} and SOLOv2~\cite{wang2020solov2} by taking them as the base architectures for our method.

\subsection{Weakly supervised segmentation}

\textbf{Weakly supervised semantic segmentation.}
A number of methods have been proposed to learn semantic segmentation with image-level class labels~\cite{durand2017wildcat,jin2017webly,ahn2018learning,fan2018associating,sun2020mining}, points~\cite{bearman2016s,qian2019weakly}, scribbles~\cite{lin2016scribblesup,vernaza2017learning,tang2018normalized,tang2018regularized} and bounding boxes~\cite{dai2015boxsup,papandreou2015weakly,khoreva2017simple,song2019box,kulharia2020box2seg}. Among them, box-supervised semantic segmentation is probably most related, and recent methods such as Box2Seg~\cite{kulharia2020box2seg} have achieved impressive performance on Pascal VOC~\cite{everingham2010pascal}. These methods often use MCG~\cite{arbelaez2014multiscale} and GrabCut~\cite{rother2004grabcut} to obtain segmentation pseudo-labels for supervising subsequent tasks. However, they focus on semantic segmentation which does not distinguish different object instances.

\textbf{Weakly supervised instance segmentation.}
Here, the term ``weakly supervised'' can either refer to the relaxed supervision on bounding box location~\cite{zhou2018weakly,ahn2019weakly}, or the absence of mask annotations~\cite{khoreva2017simple,hsu2019weakly,arun2020weakly,tian2021boxinst}. The former can be viewed as an extension of weakly supervised object detection~\cite{bilen2016weakly}, whereas our work falls into the second category. Among the latter methods, Hsu et al.~\cite{hsu2019weakly} leverages the fact that bounding boxes tightly enclose the objects, and proposes multiple instance learning framework based on this tightness prior. A pairwise loss is also imposed to maintain object integrity. However, their pairwise consistency is defined on all neighboring pixel pairs without distinguishing the pairwise pixel contrast. Arun et al.~\cite{arun2020weakly} proposes an annotation consistency framework which can handle weakly supervised instance segmentation with both image-level and bounding box labels. On COCO, the gap to supervised methods has remained large until recently BoxInst~\cite{tian2021boxinst} reduced this gap significantly. \textsc{DiscoBox} outperforms these methods while additionally targeting semantic correspondence.

\subsection{Finding correspondence}

\textbf{Local features.} Using local features to match the keypoints across different views has been widely used in 3D vision problems such as structure from motion. Over the past decade, the methods have evolved from hand-crafted ones~\cite{lowe2004distinctive,bay2006surf,tola2009daisy} to learning-based ones using decision tree and deep neural networks~\cite{rosten2006machine,verdie2015tilde,han2015matchnet,yi2016lift,ono2018lf,sarlin2020superglue} with extremely abundant literature. These methods primarily focus on multi-view association for the same object instance or scene, which differs from our task despite the underlying strong connection.

\textbf{Semantic correspondence.}
Semantic correspondences has been a challenging problem.
The problem probably dates back to SIFTFlow~\cite{liu2010sift} which uses hand-crafted features to establish the correspondence. More recent methods have universally resorted to deep networks as powerful feature extractors~\cite{rocco2017convolutional,min2019hyperpixel,liu2020semantic}. The challenge of this task is further aggravated by the costly nature of correspondence annotation. Existing datasets~\cite{ham2016proposal,min2019spair} are relatively small in size, and only provide sparse correspondence ground truths since manually annotating dense ones is prohibitive. In light of this challenge, weakly supervised semantic correspondence are proposed to learn correspondence without correspondence ground truths~\cite{hung2019scops,rocco2018end,chen2019show,lee2019sfnet,jeon2020guided,min2020learning}. In addition, existing benchmarks and methods have predominantly focused on ``object-centric'' scenarios where each image is occupied by a major object. In this work, we further add challenge to the task by considering a more generalized multi-object scenario with object localization in the loop.
\section{Method}

\begin{figure*}[t]
\centering
\includegraphics[width=0.97\linewidth]{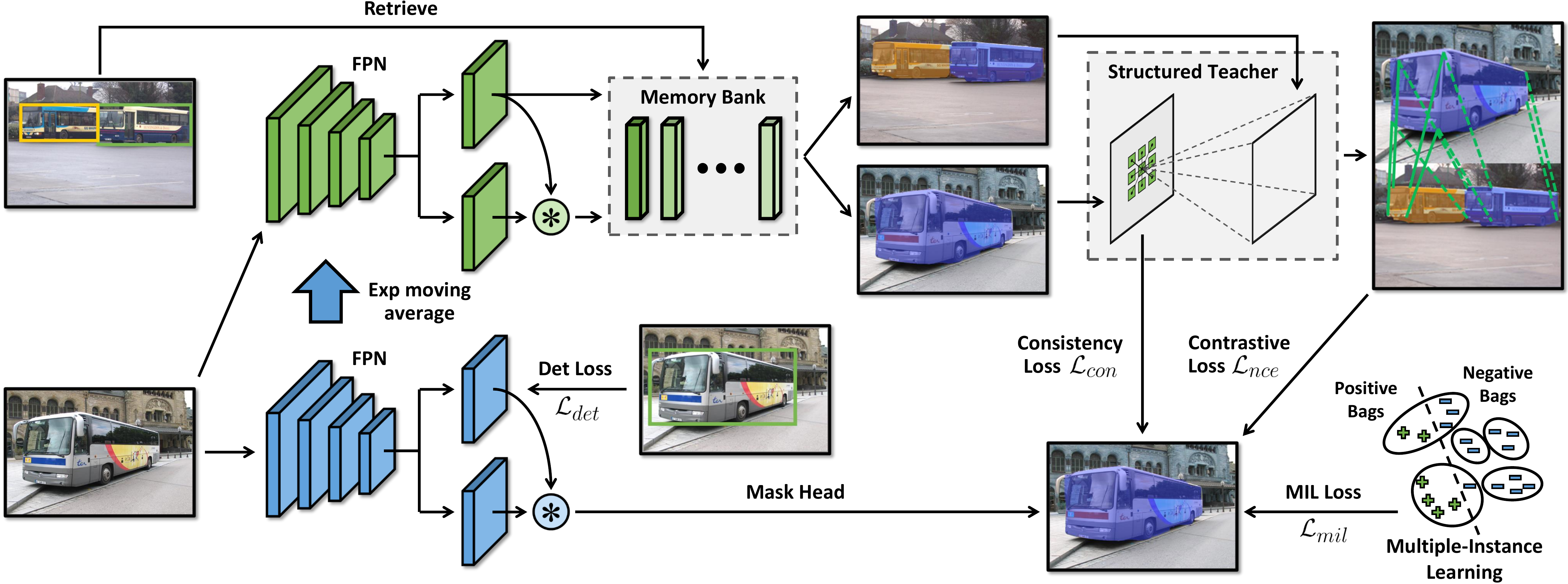}
\caption{Overview of \textsc{DiscoBox}. We design a self-ensembling framework where a structured teacher generates refined instance segmentation mask and establishes dense correspondence between intra-class box proposals to guide the task network. Best viewd in color.}
\label{fig:arch}
\vspace{-0.1cm}
\end{figure*}

We define the following notations for the variables in our problem, and use them throughout the rest of the paper. We denote the input image as $\bm{I}$. Given any instance segmentation backbone, we assume that a set of box region proposals $\bm{R}=\{\bm{r}^n|n=1,..., N\}$ are generated. Each box proposal corresponds to an RoI feature map $\bm{f}^n$ of size $C\times H \times W$. Additionally, instance segmentation produces a set of object masks $\bm{M}=\{\bm{m}^n|n=1,.., N\}$, where each $\bm{m}^n$ is an $H \times W$ probability map associated with $\bm{r}^n$. Fig.~\ref{fig:arch} illustrates an overview of the proposed framework.

\subsection{Task network}
\textsc{DiscoBox} is agnostic to the task networks. We therefore base its design on YOLACT++~\cite{bolya2020yolact++} and SOLOv2~\cite{wang2020solov2}, two recent one-stage instance segmentation frameworks.

\textbf{YOLACT++.} The architecture comprises the following components: 1) \textbf{Prediction head.} The framework adopts anchor-based one-stage detection where the prediction head outputs a set of box proposals containing the predicted coordinates and class probabilities. 2) \textbf{Mask head.} YOLACT++ proposes a PrototypeNet module to generate $D$ latent segment proposals at image level, and use the prediction head to also predict a mask coefficient (a $D$-dim vector) for every box proposal. The mask activation of each proposal is therefore a result of the weighted combination of the segment proposals and the mask coefficient. 3) \textbf{Backbone.} Feature Pyramid Network (FPN)~\cite{lin2017feature} is adopted as the backbone, where the pyramid features are scaled up by fusing the skip connected backbone features with higher resolutions.

\textbf{SOLOv2.}
We also consider an alternative design based on SOLOv2~\cite{wang2020solov2}. SOLOv2 is a state-of-the-art one-stage framework which directly predicts instance masks in a box-free, grouping-free and fully convolutional manner. This is done by decoupling the object mask generation into a mask kernel prediction and mask feature learning, together with a parallelizable matrix non-maximum suppression algorithm. SOLOv2 also adopts FPN as the backbone, where the mask kernel is predicted at each pyramid level and a unified mask feature is obtained at 1/4 scale. A slight difference with our YOLACT++ based framework is that no box proposals are directly predicted. We therefore take the tightly enclosed box from each mask as our box proposal for cropping $\bm{f}^n$.

Our framework is based on the original designs and implementations of YOLACT++ and SOLOv2, while Fig.~\ref{fig:arch} gives an abstracted illustration. We kindly ask the readers to refer to~\cite{bolya2020yolact++,wang2020solov2} for more details. We follow the same classification- and box-related training losses which we will jointly term $\mathcal{L}_{det}$ in our paper. This involves $\mathcal{L}_{cls}+\mathcal{L}_{box}$ in YOLACT++, and $\mathcal{L}_{cate}$ in SOLOv2. Since mask annotations are not available, we replace $\mathcal{L}_{mask}$ with the following multiple instance learning (MIL) loss.

\textbf{Multiple instance learning (MIL).}
MIL allows one to weakly supervise a task with inexact labels. We follow~\cite{hsu2019weakly} which proposes an MIL framework leveraging the bounding box tightness prior. Given a box which tightly encloses an object, every row and column contains at least one foreground pixel and can be treated as positive bags. Negative bags can be similarly constructed if the rows and columns have zero overlap with the ground truth box. Denote $\bm{b}_i$ the set of mask probabilities of the pixel instances belonging to bag $i$ of $\bm{r}^n$, the MIL loss for YOLACT++ is defined as:
\begin{displaymath}
\mathcal{L}_{mil} = -\sum_{i} y_i \log(\max \bm{b}_i) + (1 - y_i) \log(1 - \max \bm{b}_i)
\end{displaymath}
where $y_i = 1$ if bag $i$ is positive, and $y_i = 0$ otherwise. For SOLOv2, $\mathcal{L}_{mil}$ is similarly defined with Dice Loss~\cite{wang2020solov2}.

\subsection{Structured teacher.}

The produced segmentation from MIL is still coarse in general. Our main idea is to consider self-ensembling~\cite{tarvainen2017mean} which imposes self-consistency between perturbed models as a self-supervision to improve the representation. Self-ensembling has been a key contributor to the recent success of semi-supervised learning~\cite{sohn2020fixmatch,xie2020self}. But unlike these methods which often use augmentation and random dropout to create a noisy student, our problem allows us to form a powerful perturbed teacher through modeling the structured relationships. Promoting contrast sensitive smoothness has been an important structured inductive bias in segmentation~\cite{krahenbuhl2011efficient}. Instead of achieving this in one shot through post processing~\cite{hsu2019weakly}, our key motivation is to guide the representation with structured inductive bias in a more gradual manner with a mean-field perturbed teacher.

We define a random field $\bm{X} = \{\bm{X}^n|n=1,...,N\}$ over a graph $\bm{G} = (\bm{V}, \bm{E})$ where $\bm{x}^n\in \{0, 1\}^{H \times W}$ is a labeling of $\bm{X}^n$ in the box proposal $\bm{r}^n$. Each node $\bm{v}_i, i\in \bm{r}^n$ from box $a$ is sparsely connected with its 8 immediate neighboring nodes $\{\bm{v}_j|j \in \bm{N}_p(i)\}$, and densely connected with all the nodes $\{\bm{v}_k|k \in \bm{r}^s, s\in \bm{N}_c(n)\}$ from another intra-class box $s$. We then define the following Gibbs energy:
\begin{displaymath}
E(\bm{x}^n,\bm{T}_{ns}) = \tau_u(\bm{x}^n) + \tau_p(\bm{x}^{n}) + \sum_{\mathclap{s\in \bm{N}_c(n)}} \tau_c(\bm{x}^n, \bm{T}_{ns})
\end{displaymath}
where $\tau_u(\bm{x}^n)=\sum_i\psi(x_i^n)$ are the unary potentials taking the initial output $\bm{m}^n$ from the instance segmentation head. $\tau_p(\bm{x}^{n})$ are the pairwise potentials defined as:
\begin{displaymath}
\tau_p(\bm{x}^{n})=~\sum_{\mathclap{i\in\bm{r}^n,j\in \bm{N}_p(i)}}~~w_1\exp\Big(-\frac{|\bm{I}_{i}^n - \bm{I}_j^n|^2}{2\zeta^2}\Big)[x_i^n \neq x_j^n]
\end{displaymath}
where $\bm{I}_{i}^n$ and $\bm{I}_{j}^n$ are the RGB colors of pixel $i$ and $j$ in box $n$, and $[x_i^n \neq x_j^n]$ is a label compatibility function given by the Potts model. Finally, $\tau_c(\bm{x}^{n}, \bm{x}^s, \bm{T}_{ns})$ are the cross-image potentials which simultaneously models the dense correspondence $\bm{T}_{ns}$ and the cross-image pairwise labeling relationship. The term is defined with the following energy:
\begin{displaymath}
\begin{split}
&\tau_c(\bm{x}^{n}, \bm{T}_{ns}) = \\
&-w_2\sum_{\mathclap{i\in\bm{r}^n,k\in\bm{r}^s}} \bm{T}_{ns}(i,k) (C_u(i,k) + C_g(i,k))[x_i^n = x_k^s]\\
\end{split}
\end{displaymath}
where $\bm{T}_{ns}$ is a soft assignment matrix of size $HW \times HW$ between box proposals $\bm{r}^n$ and $\bm{r}^s$. In addition, $C_u(i,k)$ is a cost volume matrix which models the appearance similarity:
\begin{displaymath}
C_u(i,k) = \frac{\bm{f}_i^n \cdot \bm{f}_k^s}{|\bm{f}_i^n|\,|\bm{f}_k^s|}
\end{displaymath}
where $\bm{f}^n_i$ and $\bm{f}^s_k$ represent the RoI features of pixel $i$ in $\bm{r}^n$ and pixel $k$ in $\bm{r}^s$. And $C_g(i,k)$ is further defined as a pairwise smoothness regularization term aiming to impose geometric consistency:
\begin{displaymath}
\begin{split}
C_g(i,k) =
~\sum_{\mathclap{j \in \bm{r}^n, l \in \bm{r}^s}}~~\exp\Big(-\frac{| \mathsf{off}_{i,k} - \mathsf{off}_{j,l}|^2}{2*\gamma}\Big) \bm{T}_{ns}(j,l)
\end{split}
\end{displaymath}
where $\mathsf{off}_{i,k}$ represents the relative spatial offset between the pixel $i$ in $\bm{r}^n$ and pixel $k$ in $\bm{r}^s$. The intuition is to smooth the pairwise offsets to avoid spurious correspondence. 

\subsection{Inference}
We minimize the energy $E(\bm{x}^n,\bm{T}_{ns})$ with $\bm{x}^n$ and $\bm{T}_{ns}$ alternatively. Although the original $\tau_c(\bm{x}^{n}, \bm{T}_{ns})$ contains a different label compatibility function, its inference with $\bm{x}^n$ is exactly equivalent to the following energy:
\begin{displaymath}
\begin{split}
&\tau^*_c(\bm{x}^{n}, \bm{T}_{ns}) = \\
&\sum_{i,k} \bm{T}_{ns}(i,k) (C_u(i,k) + C_g(i,k))[x_i^n \neq x_k^s]\\
\end{split}
\end{displaymath}
Thus, $E(\bm{x}^n,\bm{T}_{ns})$ can be minimized via standard mean-field. Kindly refer to Appendix~\ref{sec:meanfield} for more details.

When fixing $\bm{x}$, we optimize $\bm{T}$ by solving an optimal transport problem~\cite{liu2020semantic} with the following energy:
\begin{displaymath}
\begin{split}
&\min_{\bm{T}_{ns}}~\tau_c(\bm{x}^{n}, \bm{T}_{ns})\\
&~~\text{s.t.}~\bm{T}_{ns} \bm{1}_{HW} = \mathbf{\mu}_n,~\bm{T}_{ns}^\top \bm{1}_{HW} = \mathbf{\mu}_s
\end{split}
\end{displaymath}
where $\mathbf{\mu}_n$, $\mathbf{\mu}_s$ represent the pixel-wise importance in $\bm{r}^n$ and $\bm{r}^s$, obtained by applying a step function to $\bm{m}^n$ and $\bm{m}^s$.

One may use differentiable Hungarian (denoted $\mathcal{H}$) such as Sinkhorn's algorithm to solve an optimal transport problem. However, with the pairwise term, directly solving becomes very hard. We therefore approximate with iterated conditional mode where $\bm{T}_{ns}$ is optimized iteratively:
\begin{itemize}[leftmargin=0cm]
    \item[] \noindent{\textbf{Initialize:}} $C_u \xleftarrow[]{} \frac{\bm{f}^n \cdot \bm{f}^s}{|\bm{f}^n|\,|\bm{f}^s|}$, $C^{0} \xleftarrow[]{} C_u(i,k)$
    \item[] \noindent{\textbf{Assign:}} $~~~~\bm{T}_{ns}^t \xleftarrow[]{} \mathcal{H}(C^t)$
    \item[] \noindent{\textbf{Update:}} $~~~C_g^{t}(i,k) \xleftarrow[]{}
    \sum_{j,l} \exp(-\frac{|\mathsf{off}_{i,k} - \mathsf{off}_{j,l}|^2}{2*\gamma}) \bm{T}_{ns}(j,l)^{t}$\\ 
 $~~~~~~~~~~~~~~~~~~~C^{t+1}(i,k) = C_u(i,k) + C_g^t(i,k)$
\end{itemize}
An illustration of the above algorithm is shown in Fig.~\ref{fig:otl}. Additional algorithm details can be found in Appendix~\ref{sec:sinkhorn}.

\begin{figure}[t]
\centering
\includegraphics[width=1.0\linewidth]{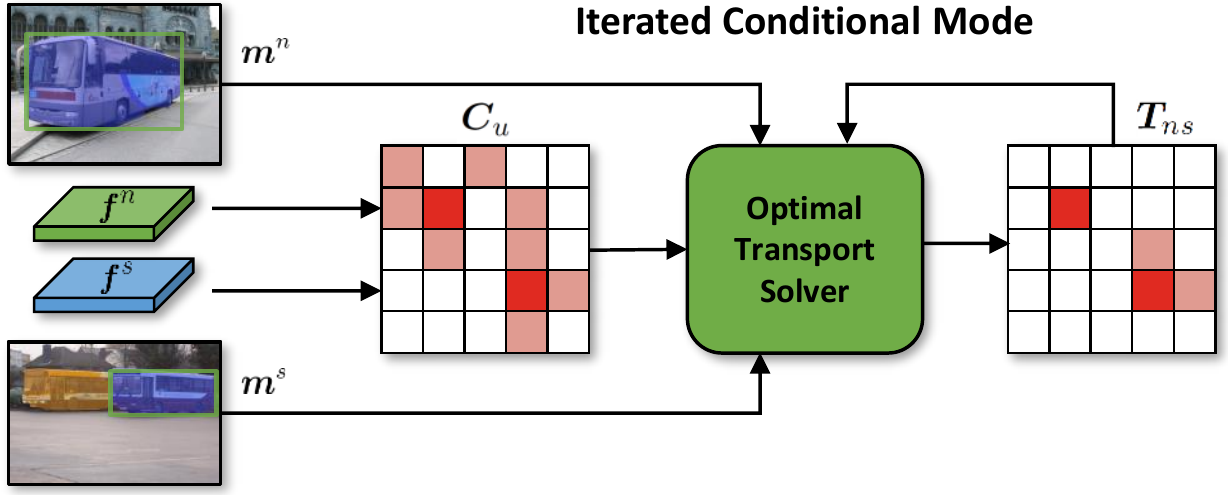}
\caption{Illustration of the proposed iterated conditional mode algorithm. Given a pair of RoI features, we use them to initialize the cost volume $\bm{C}_u$. Differentiable Hungarian is then iteratively applied to optimize $\bm{T}_{ns}$ and update $\bm{C}_g$.}
\label{fig:otl}
\end{figure}

\subsection{Learning}

With the $\bm{x}$ and $\bm{T}$ inferred from teacher, we can define the following self-ensembling losses. We impose self-consistency between our task network and teacher:
\begin{displaymath}
\begin{small}
\mathcal{L}_{con} = \frac{1}{|\bm{r}^n|} \sum_{i \in \bm{r}^n} \Big[x_i^n\log(m_i^n) +(1-x_i^n)\log(1-m_i^n)\Big]
\end{small}
\end{displaymath}
We also use the dense correspondence $\bm{T}_{ns}$ to obtain positive and negative pairs for dense contrastive learning:
\begin{displaymath}
\mathcal{L}_{nce} = \frac{1}{|\bm{r}^n|}\sum_{i \in \bm{r}^n}\log\frac{\exp(C_u(i, t_i)/\tau)}{\sum_{k\in\bm{r}^s} \exp(C_u(i,k)/\tau)}
\end{displaymath}
where $t_{i} = \arg\max_k \bm{T}_{ns}(i,k)$ and $\tau$ is the temperature. Our joint training loss therefore can be written as:
\begin{displaymath}
    \mathcal{L} = \mathcal{L}_{det} + \alpha_{mil} \mathcal{L}_{mil} + \alpha_{con} \mathcal{L}_{con} + \alpha_{nce} \mathcal{L}_{nce}
\end{displaymath}

\subsection{Exponential moving averaged teacher}
To further strengthen the teacher with model-level augmentation, improved stability, and better consistency between iterations, we follow~\cite{tarvainen2017mean,he2020momentum} to obtain a mean teacher with exponential moving average. This is done by maintaining another network sharing the same architecture with the task network and update the parameters following exponential moving average:
\begin{displaymath}
    \theta_{t} \xleftarrow[]{} m \theta_t + (1-m) \theta_s
\end{displaymath}
where $\theta_t$, $\theta_s$ are the parameters of teacher network and task network. $m$ is the momentum and is set to 0.999 following~\cite{he2020momentum}. We do not train the teacher and only update it.

\subsection{Object retrieval with memory bank}

To conveniently obtain the object pairs for semantic correspondence, we construct a first-in-first-out (FIFO) queue for each category, where we push the RoI features $\bm{f}$ and masks $\bm{m}$ from each batch. This allows us to reuse the RoI features and masks and construct object pairs without much extra computation. During training, the model will retrieve similar intra-class objects from the object bank. After all losses are computed, we push all the objects except the object with area  $< 32 \times 32 $ into the object banks. Inter-class objects are stored in different queues. Only intra-class objects share the same object bank. In practice, we set the size of object bank of one category as 100.
\section{Experiments}

We conduct experiments on 4 datasets: PASCAL VOC 2012 (VOC12)~\cite{everingham2010pascal}, COCO~\cite{lin2014microsoft}, PF-PASCAL~\cite{ham2016proposal}, PASCAL 3D+~\cite{xiang2014beyond}. 
We test instance segmentation on VOC12 and COCO, and semantic correspondence on the other two.

\subsection{Datasets and metrics}

\textbf{COCO.} COCO contains 80 semantic categories. We follow the standard partition which includes train2017 (115K images) and val2017 (5K images) for training and validation. We also report our results on the test-dev split. During training, we only use the box annotations.

\textbf{VOC12.} VOC12 consists of 20 categories with a training set of around 10,500 images and a validation set of around 5,000 images. Around 1,500 images of the validation set contain the instance segmentation annotations.

\textbf{PF-PASCAL.} The PF-PASCAL dataset contains a selected subset of object-centric images from PASCAL VOC. It contains around 1,300 image pairs with 700 pairs for the training set and 300 pairs for the validation set, and 300 image pairs for the test sets respectively. There is only one conspicuous object in the middle of the image. Each image pair contain two intra-class objects.

\textbf{PASCAL 3D+.} PASCAL 3D+ contains the annotations of object poses, landmarks and 3D CAD models in addition to bounding boxes, and consists of 12 rigid categories where each has 3,000 object instances on average. We evaluate multi-object correspondence on PASCAL 3D+ dataset. The availability of both bounding boxes and landmarks, as well as other 3D information makes it an ideal dataset to evaluate multi-object semantic correspondence. We construct the benchmark on the 12 rigid categories of PASCAL 3D+ and follow the official VOC12 partitioning of the validation set, where images only containing the 8 non-rigid classes are removed. For training, we still preserve the full VOC12 training set and annotations (20 classes).

As PASCAL 3D+ does not provide image pairs, we need to generate image pairs and keypoint pairs on PASCAL for the correspondence evaluation. We enumerate all pairwise combinations of two images on the PASCAL 3D+ validation set. For any pairwise images, if both contain at least one intra-class object in common, we mark them as matched and keep this pair for evaluation. The second step is to generate the sparse correspondence ground truths on top of the matched image pairs using the provided keypoints. For any pairwise images, we find all combinations of intra-class object pairs and use the keypoint pairs between these object pairs as the correspondence ground-truth. Due to occlusion, some keypoints may be missing and are ignored during the evaluation. Note that we also ignore any pairwise objects where the difference between their 3D orientations is greater than 60 degrees, since a large orientation gap often results in very few valid keypoint pairs.

\textbf{Multi-object correspondence metric.} Similar to object detection, we introduce a precision-recall based metric with average precision (AP). We assume that there is a confidence associated with each predicted correspondence, and we define it as the multiplication of the pairwise box confidence in this work. This allows us to compute precision and recall by defining true positive ($\text{TP}$), false positive ($\text{FP}$) and false negative ($\text{FN}$). Since PASCAL 3D+ only provides sparse correspondence ground truths, the challenge here is to correctly ignore some of the correspondence predictions that are far away from any ground truth but are correct. To this end, we follow a keypoint transfer setting where we always define a \textit{source} side $s$ and a \textit{target} side $t$ for any pairwise objects. Given a ground truth $(\bm{g}_j^s, \bm{g}_j^t)$, a predicted correspondence $(\bm{p}_i^s, \bm{p}_i^t)$ and a distance threshold $\alpha$:

\begin{small}
\vspace{-0.2cm}
\begin{displaymath}
\begin{split}
    \text{TP}_i &= \frac{\sum_j \mathbbm{1}[|\bm{p}_i^s - \bm{g}_j^s| \leq \alpha] \times \mathbbm{1}[|\bm{p}_i^t - \bm{g}_j^t| \leq \alpha]}{\sum_j \mathbbm{1}[|\bm{p}_i^s - \bm{g}_j^s| \leq \alpha] + \mathbbm{1}[\sum_j \mathbbm{1}[|\bm{p}_i^s - \bm{g}_j^s| \leq \alpha]=0]}\\[6pt]
    \text{FP}_i &= \frac{\sum_j \mathbbm{1}[|\bm{p}_i^s - \bm{g}_j^s| \leq \alpha] \times \mathbbm{1}[|\bm{p}_i^t - \bm{g}_j^t| > \alpha]}{\sum_j \mathbbm{1}[|\bm{p}_i^s - \bm{g}_j^s| \leq \alpha] + \mathbbm{1}[\sum_j \mathbbm{1}[|\bm{p}_i^s - \bm{g}_j^s| \leq \alpha]=0]}\\[6pt]
    \text{FN}_i &= \left\{
    \begin{array}{ll}
    1 & \mbox{if~$~\sum_j \mathbbm{1}[|\bm{p}_i^s - \bm{g}_j^s| \leq \alpha]=0$},\\
    0 &\mbox{otherwise}
    \end{array}
    \right.
\end{split}
\end{displaymath}
\end{small}\\
We term the average precision as $\text{AP}@\alpha$ where $\alpha$ is a threshold relative to the box diagonal. We then define the final AP as: $\text{mean}(\text{AP}@\{0.75\%,1\%,1.5\%,2\%,3\%\})$.

\begin{table*}[t]
\centering
\resizebox{\textwidth}{!}{
\addtolength{\tabcolsep}{3pt}
\begin{tabular}{lccccccccccc}
\toprule
Method & Backbone & Type & Arch  & Sup & FPS & $\text{AP}$ & $\text{AP}_{\text{50}}$ & $\text{AP}_{\text{75}}$ & $\text{AP}_{\text{S}}$ & $\text{AP}_{\text{M}}$ & $\text{AP}_{\text{L}}$ \\
\midrule
MNC~\cite{dai2016instance} & ResNet-101 & Two-stage & MNC & Mask & $<$2.8 & 24.6 & 44.3 & 24.8 & 4.7 & 25.9 & 43.6 \\
FCIS~\cite{li2017fully} & ResNet-101 & Two-stage & FCIS & Mask & 6.6 &  29.2 & 49.5 & - & 7.1 & 31.3 & 50.0\\
Mask R-CNN~\cite{he2017mask} &  ResNet-101 & Two-stage & Mask R-CNN & Mask & 5 &  35.7 & 58.0 & 37.8 & 15.5 & 38.1 & 52.4 \\
Mask R-CNN~\cite{he2017mask} &  ResNeXt-101 & Two-stage & Mask R-CNN & Mask & $<$5 &  37.1 & 60.0 & 39.4 & 16.9 & 38.9 & 53.5 \\
YOLACT-550~\cite{bolya2019yolact} &  ResNet-50 & One-stage & YOLACT & Mask & \textbf{45} & 28.2 & 46.6 & 29.2 & 9.2 & 29.3 & 44.8 \\
YOLACT-700~\cite{bolya2019yolact} &  ResNet-101 & One-stage & YOLACT & Mask & 23.4 & 31.2 & 50.6 & 32.8 & 12.1 & 33.3 & 47.1 \\
PolarMask~\cite{xie2020polarmask} &  ResNet-101 & One-stage & PolarMask & Mask & 12.3 & 32.1 & 53.7 & 33.1 & 14.7 & 33.8 & 45.3 \\
YOLACT-550++~\cite{bolya2020yolact++} & ResNet-50-DCN & One-stage & YOLACT++ & Mask & 33.5 & 34.1 & 53.3 & 36.2 & 11.7 & 36.1 & 53.6 \\
YOLACT-550++~\cite{bolya2020yolact++} & ResNet-101-DCN & One-stage & YOLACT++ & Mask & 27.3 & 34.6 & 53.8 & 36.9 & 11.9 & 36.8 & 55.1 \\
SOLOv2~\cite{wang2020solov2} & ResNet-101-DCN & One-stage & SOLOv2 & Mask & 10.3 & \textbf{41.7} & \textbf{63.2} & \textbf{45.1} & \textbf{18.0} & \textbf{45.0} & \textbf{61.6} \\
\midrule
BBTP$\dagger$~\cite{hsu2019weakly} & ResNet-101 & Two-stgae & Mask R-CNN  & Box & $<$5 & 21.1 &  45.5  & 17.2   & 11.2 & 22.0 & 29.8\\
BoxInst~\cite{tian2021boxinst} & ResNet-101-DCN & One-stage & CondInst~\cite{tian2020conditional} & Box & - & 35.0 & 59.3 & 35.6 & 17.1 &37.2 & 48.9 \\
\textsc{DiscoBox} (Ours)$\dagger$ & ResNet-50-DCN & One-stage & YOLACT++ & Box & \textbf{34.5} & 26.9 & 48.6 & 26.3 & 9.6 & 27.8 & 42.1 \\
\textsc{DiscoBox} (Ours)$\dagger$ & ResNet-50 & One-stage & SOLOv2 & Box & 18.5 & 31.4 & 52.6 & 32.2 & 11.5 & 33.8 & 50.1 \\
\textsc{DiscoBox} (Ours) & ResNet-50 & One-stage & SOLOv2 & Box & 18.5 & 32.0 & 53.6 & 32.6 & 11.7 & 33.7 & 48.4 \\
\textsc{DiscoBox} (Ours) & ResNet-101-DCN & One-stage & SOLOv2 & Box & 10.3 & 35.8 & 59.8 & 36.4
& 16.9 & 38.7 & 52.1 \\
\textsc{DiscoBox} (Ours) & ResNeXt-101-DCN & One-stage & SOLOv2 & Box & 7.4 & \textbf{37.9} & \textbf{61.4} & \textbf{40.0} & \textbf{18.0} & \textbf{41.1} & \textbf{53.9} \\
\bottomrule
\end{tabular}
}
\vspace{0.05cm}
\caption{Main results on COCO. $\dagger$ indicates that the results are on the COCO validation 2017 split. The rest results are on COCO test-dev. \textsc{DiscoBox} with SOLOv2/ResNet-50 outperforms BBTP~\cite{hsu2019weakly} by 10.3\% on COCO validation 2017. Our best model achieves 37.9\% mAP on test-dev, which outperforms some competitive supervised methods such as Mask R-CNN in absolute performance.
}
\label{tab:coco_exp}
\end{table*}

\begin{table}[t]
\centering
\resizebox{\linewidth}{!}{
\begin{tabular}{lcccccccc}
\toprule
\textbf{Method} & \textbf{Backbone} & \textbf{Arch} & $\text{AP}_{\text{25}}$ & $\text{AP}_{\text{50}}$ & $\text{AP}_{\text{70}}$ & $\text{AP}_{\text{75}}$ \\
\midrule
SDI~\cite{khoreva2017simple} & VGG-16 & DeepLabv2 & - & 44.8 & - & 16.3 \\
BBTP~\cite{hsu2019weakly} & ResNet-101 & Mask R-CNN & 75.0 & 58.9 & 30.4 & 21.6 \\
Arun et al.~\cite{arun2020weakly} & ResNet-101 & Mask R-CNN & 73.1 & 57.7 & 33.5 & 31.2 \\
BoxInst~\cite{tian2021boxinst} & ResNet-50 & CondInst &- & 59.1 & - & 34.2 \\
BoxInst~\cite{tian2021boxinst} & ResNet-101 & CondInst & - & 61.4 & - & 37.0 \\
\textsc{DiscoBox} & ResNet-50-DCN & YOLACT++ & \textbf{75.2}& \textbf{63.6} & 41.6 & 34.1 \\
\textsc{DiscoBox} & ResNet-50 & SOLOv2 & 71.4 & 59.8 & 41.7 & 35.5 \\
\textsc{DiscoBox} & ResNet-101 & SOLOv2 & 72.8 & 62.2 & \textbf{45.5} & \textbf{37.5} \\
\bottomrule
\end{tabular}
}
\vspace{0.05cm}
\caption{Main results on the VOC12 validation set. \textsc{DiscoBox} outperforms all previous methods with state-of-the-art results.
}
\label{tab:voc_exp}
\end{table}

\subsection{Implementation details} \label{sec:exp_impl}

\textbf{Training.} We use stochastic gradient descent (SGD) for network optimization. For loss weights, we set $\alpha_{mil}$, $\alpha_{con}$,  $\alpha_{nce}$ as $10, 2, 0.1$ on YOLACT++ and set $\alpha_{mil}$, $\alpha_{con}$, $\alpha_{nce}$ as $1, 1, 0.1$ on SOLOv2. Kindly refer to Appendix~\ref{sec:add_impl} for additional implementation details.

\subsection{Weakly supervised instance segmentation} \label{sec:wssc}

\begin{figure*}[t]
\centering
\includegraphics[width=\linewidth]{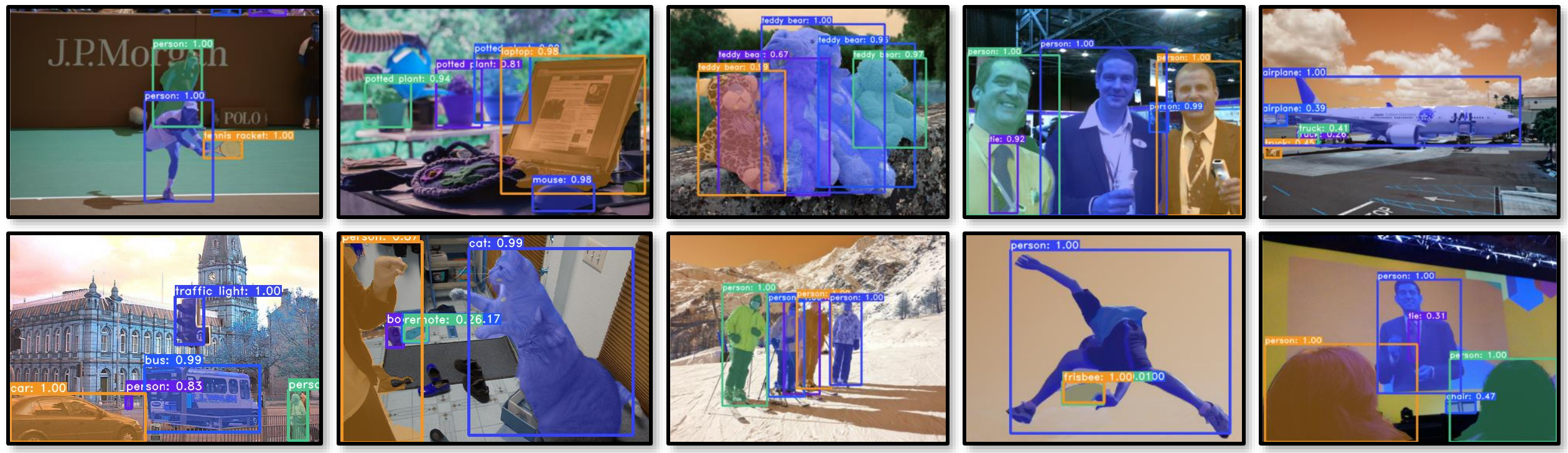}\\
\caption{Visualization of instance segmentation on COCO (YOLACT++/ResNet-50-DCN).}
\label{fig:vis:inst}
\vspace{-0.15cm}
\end{figure*}

\textbf{Main results.} We evaluate instance segmentation on COCO and VOC12, with the main results reported in Tab.~\ref{tab:coco_exp} and~\ref{tab:voc_exp}, respectively. \textsc{DiscoBox} outperforms BBTP~\cite{hsu2019weakly} by $10.3\%$ mAP on the COCO validation 2017 split with a smaller backbone (ResNet-50). \textsc{DiscoBox} also outperforms BoxInst~\cite{tian2021boxinst} which is the current state-of-the-art box-supervised method on both COCO and VOC12. Notably, BoxInst/ResNet-101-DCN also adopts BiFPN~\cite{tan2020efficientdet}, an improved variant of FPN~\cite{lin2017feature}. Fig.~\ref{fig:vis:inst} and Appendix~\ref{sec:add_vis} additionally visualize the instance segmentation results.

\begin{figure*}[t]
\centering
\includegraphics[width=\linewidth]{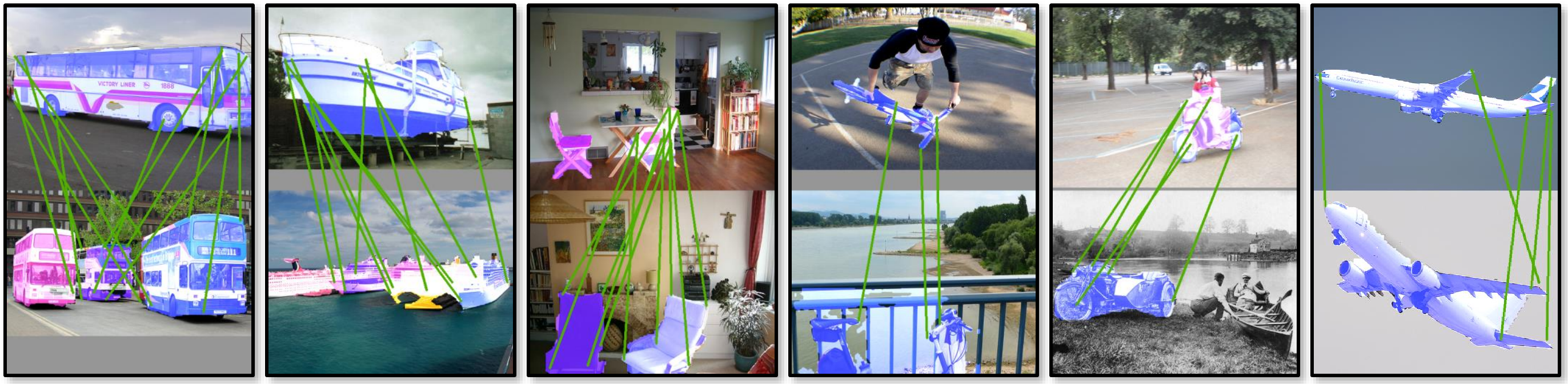}
\caption{Visualization of multi-object semantic correspondence on PASCAL 3D+ (YOLACT++/ResNet-50-DCN).}
\label{fig:vis:corr}
\end{figure*}

\begin{figure}[t]
\vspace{-0.1cm}
\centering
\includegraphics[width=1.0\linewidth]{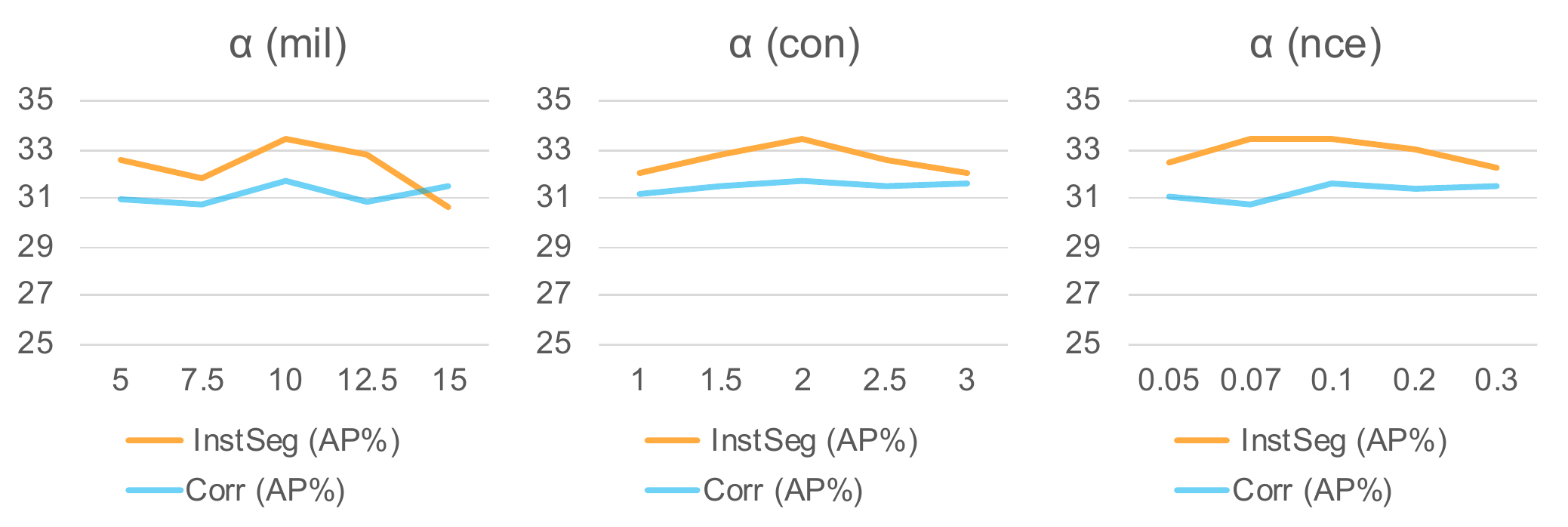}
\caption{Sensitivity analysis with the loss weights on instance segmentation (VOC12) and semantic correspondence (PASCAL 3D+). Analysis conducted on YOLACT++/ResNet-50-DCN.}
\label{fig:alpha}
\end{figure}

\textbf{Analysis.} We perform ablation study on VOC12 with $\mathcal{L}_{mil}$, $\mathcal{L}_{con}$ and $\mathcal{L}_{nce}$. The results in Tab.~\ref{tab:ablation} show consistent improvements from $\mathcal{L}_{con}$ and $\mathcal{L}_{nce}$, demonstrating the benefit of the structured teacher. We also conduct sensitivity analysis with the loss weights on both instance segmentation (VOC12)\footnote{Here, the AP follows COCO evaluation: mean(AP@\{50,55,...,95\}).}
and semantic correspondence (PASCAL 3D+, see Sec.~\ref{sec:exp_corr}). The results in Fig.~\ref{fig:alpha} show that \textsc{DiscoBox} is not sensitive to weight changes.

\begin{table}[t]
\centering
\resizebox{\linewidth}{!}{
\addtolength{\tabcolsep}{1pt}
\begin{tabular}{ccccccccccc}
\toprule
$\mathcal{L}_{mil}$ & $\mathcal{L}_{con}$ & $\mathcal{L}_{nce}$ &  $\text{AP}_{\text{50}}^{\,\text{y}}$ & $\text{AP}_{\text{70}}^{\,\text{y}}$ & $\text{AP}_{\text{75}}^{\,\text{y}}$ & $\text{AP}_{\text{50}}^{\,\text{s}}$ & $\text{AP}_{\text{70}}^{\,\text{s}}$ & $\text{AP}_{\text{75}}^{\,\text{s}}$ \\
\midrule
\checkmark & - & - & 43.3 & 18.3 & 17.0 & 42.1 & 18.0 & 17.3 \\ 
\checkmark & \checkmark & -  & 62.0 & 40.1 & 33.5 & 58.1 & 40.9 & 34.9  \\ 
\checkmark & \checkmark & \checkmark & \textbf{63.6} & \textbf{41.6} & \textbf{34.1} & \textbf{59.8} & \textbf{41.7} & \textbf{35.5}\\
\bottomrule
\end{tabular}
}
\vspace{0.05cm}
\caption{Ablation study on VOC12, where `y' denotes the results obtained by YOLOCT++/ResNet-50-DCN and `s' denotes the results obtained by SOLOv2/ResNet-50.}
\label{tab:ablation}
\end{table}

\begin{table}[]
\centering
\resizebox{\linewidth}{!}{
\begin{tabular}{lcccc}
\hline
Signal & Methods  & PCK@0.05 & PCK@0.1 & PCK@0.15 \\
\hline
none & PF$_{\text{HOG}}$~\cite{han2017scnet} & 31.4 & 62.5 & 79.5 \\
\hline
image-level & WeakAlign~\cite{rocco2018end} & 49.0 & 74.8 & 84.0 \\
image-level & RTNs~\cite{kim2018recurrent} & 55.2 & 75.9 & 85.2 \\
image-level & NC-Net~\cite{rocco2018neighbourhood} & 54.3 & 78.9 & 86.0 \\
image-level & DCC-Net~\cite{huang2019dynamic} & 55.6 & 82.3 & 90.5 \\
\midrule
mask & SF-Net~\cite{lee2019sfnet} & 53.6 & 81.9 & 90.6 \\
mask & DHPF~\cite{min2020learning} & 56.1 & 82.1 & 91.1 \\
\midrule
box & \textsc{DiscoBox} & $\textbf{59.3}$ & $\textbf{87.2}$ & $\textbf{95.3}$ \\
\bottomrule
\end{tabular}
}
\vspace{0.05cm}
\caption{Results on PF-PASCAL. \textsc{DiscoBox} outperforms previous state-of-the-art methods on weakly supervised semantic correspondence without bells-and-whistles.}
\label{tab:pf-pascal}
\end{table}

\subsection{Weakly supervised semantic correspondence} \label{sec:exp_corr}

\textbf{PF-PASCAL (Object-Centric).} We first evaluate \textsc{DiscoBox} on PF-PASCAL~\cite{ham2016proposal} using YOLACT++/ResNet-50-DCN, with the main results presented in Tab.~\ref{tab:pf-pascal}. We do not  directly train the \textsc{DiscoBox} model on PF-PASCAL. Instead, we train it on the VOC12 training set, excluding those images that are present in the PF-PASCAL validation set. It is worth noting that many existing semantic correspondence methods can not be similarly trained on VOC12 without major changes, even though some of them do consider certain level of localization information such as attention. During inference, we use instance segmentation to obtain object masks, and use the structured teacher to produce dense pixel-wise correspondence by taking the masks as input. Our approach outperforms the previous weakly supervised semantic correspondence approaches with considerable margins. Such improvement can be attributed to three main factors: 1) The improved design of structured teacher which renders good correspondence quality at object-level. 2) The box-supervised learning framework which makes it possible to scale up the training using more data and obtain improved correspondence representation. 3) The high quality object localization as a result of the coupled learning framework that help to guide the correspondence.

\textbf{PASCAL 3D+ (Multi-Object).} Finally, we benchmark \textsc{DiscoBox} and several baselines on PASCAL 3D+. Tab.~\ref{tab:nocc} lists the main results and Fig.~\ref{fig:vis:corr} visualizes some predicted correspondence. The comparing methods in Tab.~\ref{tab:nocc} are defined as follows: \textbf{Identity:} We align each pair of images only considering the positions of pixels. \textbf{SCOT:} A modified version of~\cite{liu2020semantic} by removing beam search and keeping their matching module on our RoI features. \textbf{\textsc{DiscoBox-}:} Our model trained on VOC12 without dense NCE loss, but using teacher during inference for correspondence. \textbf{\textsc{DiscoBox}:} Our full approach. We use YOLACT++/ResNet-50-DCN for all methods. Our method does not include beam search with the validation data and label~\cite{liu2020semantic}, and is therefore purely box-supervised. The results show the effectiveness of our proposed teacher and dense contrastive learning.

\begin{table}[]
\centering
\resizebox{\linewidth}{!}{
\addtolength{\tabcolsep}{2pt}
\begin{tabular}{lcccccc} 
\toprule
Methods & ~AP~ & $\!\text{AP}_{\text{0.75}}\!$ & $\,\text{AP}_{1}\,$ & $\text{AP}_{\text{1.5}}$ & $\,\text{AP}_{\text{2}}\,$ & $\,\text{AP}_{\text{3}}\,$\\ 
\midrule
Identity & 26.6 & 10.5 & 16.3 & 26.2 & 34.2 & 46.0  \\ 
SCOT~\cite{liu2020semantic} & 29.3 & 13.2 & 19.8 & 29.8 & 36.0 & 47.3  \\ 
\textsc{DiscoBox-} & 30.9 & 15.6  & 21.3 & 30.6 & 38.0 & 48.9 \\ 
\textsc{DiscoBox} & \textbf{31.7} & \textbf{15.8} & \textbf{21.4} & \textbf{31.8} & \textbf{39.5} & \textbf{50.3} \\
\bottomrule
\end{tabular}
}
\vspace{0.05cm}
\caption{Results of multi-object correspondence on PASCAL 3D+, where $\text{AP}_{\text{0.75}}$, $\text{AP}_{1}$, $\text{AP}_{\text{1.5}}$, $\text{AP}_{\text{2}}$, and $\text{AP}_{\text{3}}$ represent $\text{AP}@\alpha$ with thresholds $\alpha\in\{0.75\%, 1\%, 1.5\%, 2\%, 3\%\}$ relative to the box diagonal. AP is defined as $\text{mean}(\text{AP}_{\text{0.75}},\text{AP}_{1},\text{AP}_{\text{1.5}},\text{AP}_{\text{2}},\text{AP}_{\text{3}})$.}
\label{tab:nocc}
\vspace{-0.2cm}
\end{table}
\section{Conclusions}

We presented \textsc{DiscoBox}, a novel framework able to jointly learn instance segmentation and semantic correspondence from box supervision. Our proposed self-ensembling framework with a structured teacher has led to significant improvement with state of the art performance in both tasks. We also proposed a novel benchmark for multi-object semantic correspondence together with a principled evaluation metric. With the ability to jointly produce high quality instance segmentation and semantic correspondence from box supervision, we envision that \textsc{DiscoBox} can scale up and benefit many downstream 2D and 3D vision tasks.

\noindent\textbf{Acknowledgement:} We would like to sincerely thank Xinlong Wang, Zhi Tian, Shuaiyi Huang, Yashar Asgarieh, Jose M. Alvarez, De-An Huang, and other NVIDIA colleagues for the discussion and constructive suggestions.

{\small
\bibliographystyle{unsrt}
\bibliography{ref}
}

\newpage
\appendix

\section{Mean-field inference} \label{sec:meanfield}
Recap that we denote the input image as $\bm{I}$, and have a set of box region proposals $\bm{R}=\{\bm{r}^n|n=1,..., N\}$. Each box proposal corresponds to an RoI feature map $\bm{f}^n$ of size $C\times H \times W$. Additionally, the box proposals correspond to a set of object masks $\bm{M}=\{\bm{m}^n|n=1,.., N\}$, where each $\bm{m}^n$ is an $H \times W$ probability map.

Without loss of generality, we assume $\bm{x}^n\in \{0, 1\}^{H \times W}$ is the labeling of a box $\bm{r}^n$ from current batch, and $\bm{x}^s\in \{0, 1\}^{H \times W}$ is the labeling of an intra-class box $\bm{r}^s$, where $s\in \bm{N}_c(n)$ is an index of all the retrieved boxes. We denote $\bm{m}^n$ and $\bm{m}^s$ the predicted mask probability maps and $\bm{I}^n$ and $\bm{I}^s$ the cropped RGB images of $\bm{r}^n$ and $\bm{r}^s$. We further denote $i,j,k$ the indices of box pixels and $\bm{N}_p(i)$ the set of 8-connected immediate neighbors of pixel $i$. We also assume a dense correspondence $\bm{T}_{ns}$ has been established between $\bm{r}^n$ and $\bm{r}^s$ based on the cost volume $\bm{C} = \bm{C}_u + \bm{C}_g$. The mean field inference method is shown in Algorithm~\ref{alg:1}:

\begin{algorithm}
\caption{Mean Field Inference.}\label{euclid}
  \begin{algorithmic}[1]
    \Procedure{Mean Field}{$\bm{m}^n, \bm{m}^s, \bm{I}^n, \bm{I}^s, \bm{T}_{ns}, \bm{C}$}
      \State $\varphi(x) = \mathbbm{1}[x \leq 0.5] * 0.3 + \mathbbm{1}[x > 0.5] * 0.7$ \Comment{threshold function.}
      \State $\bm{q}^n \gets -\log(\varphi(\bm{m}^n)),~\bm{q}^s \gets -\log(\varphi(\bm{m}^s)$)
      \State $\bm{k}(i,j) \xleftarrow[]{} w_1\exp(-\frac{|\bm{I}^n_{i} - \bm{I}^n_{j}|^2}{2\zeta^2})$ \Comment{pairwise kernels}
      \While{not converge} \Comment{iterate until convergence}
        \For {$i \in \bm{r}^n$}
            \State $\hat{q}^n_{i} \gets 0$
            \For{$j \in \bm{N}(i)$} \Comment{pairwise potentials}
                \State $\hat{q}^n_i \gets \hat{q}^n_i + \bm{k}(i,j) q^n_j$
            \EndFor
            \For{$s \in \bm{N}_c(n)$}
                \For{$k \in \bm{r}^s $} \Comment{cross-image potentials}
                    \State{$\hat{\mathbf{x}}^n_{i} \gets \hat{\mathbf{x}}^n_{i} + w_2\bm{T}_{ns}(i,k) \bm{C}(i,k) \bm{x}^s_{k}$} 
                \EndFor
            \EndFor
        \EndFor
        \State $\bm{q}^n \gets \varphi(\exp(-\hat{\bm{q}}^n - \bm{q}^n))$ \Comment{local update}
        \State normalize $\bm{q}^n$ \Comment{normalization}
        \State $\bm{q}^n \gets -\log(\bm{q}^n)$
      \EndWhile\label{euclidendwhile}
      \State $\bm{x}^n \gets \mathbbm{1}[\exp(-\bm{q}^n) > 0.5]$
      \State \textbf{return} $\bm{x}^n$
    \EndProcedure
  \end{algorithmic}
  \label{alg:1}
\end{algorithm}

\section{Sinkhorn's algorithm} \label{sec:sinkhorn}

Given any cost volume $\bm{C}$, we use Sinkhorn's Algorithm ~\cite{knight2008sinkhorn} to find out the one-to-one assignment approximately. We define a threshold function $\varphi^o(x) = \mathbbm{1}[x > 0.5] * 0.4 + 0.6$, and use the function to obtain $\bm{\mu}_a = \varphi^o(\bm{m}^a)$ and $\bm{\mu}_b = \varphi^o(\bm{m}^b)$. The Sinkhorn's algorithm to obtain the transport matrix is described in the Algorithm~\ref{alg:2}:

\begin{algorithm}
  \caption{Sinkhorn's Algorithm}
  \begin{algorithmic}
  \Procedure{$\mathcal{H}$}{$\bm{C}, \bm{\mu}_a, \bm{\mu}_b, \varepsilon, t_{max}$} 
        \State $\bm{K} = e^{(-\bm{C}/\varepsilon)}$
        \State $b \gets \mathbf{1}$
        \State $t \gets 0$
        \While{$t \leq t_{max}$ \textbf{and} not converge }
            \State $\bm{a} = \frac{\mu_a}{\bm{K}\bm{b}}$
            \State $\bm{b} = \frac{\mu_b}{\bm{K}^\intercal\bm{a}}$
        \EndWhile
        \State $\bm{T} = diag(\bm(a))\bm{K}diag(\bm{b})$
        \State \textbf{return} $\bm{T}$
  \EndProcedure 
  \end{algorithmic}
  \label{alg:2}
  \vspace{0.1cm}
\end{algorithm}

\section{Additional implementation details} \label{sec:add_impl}

\subsection{Data loading and augmentation}
\textsc{Disco-Box} follows the original data loading and augmentation settings in YOLACT++ and SOLOv2:

\noindent\textbf{YOLACT++.}
In YOLACT++~\cite{bolya2020yolact++}, an input image is resized without changing its aspect ratio so that its longer side is equal to $L$. Random cropping is then applied on the resized image crop size $L \times L$. In case the crop goes outside the image, the outside part of the crop is filled with the mean RGB values. Color jittering and random flipping are then applied on top of random cropping as data augmentation. The original paper of YOLACT++ has reported results with different $L$ such as $550$ and $700$. We follow the same setting of loading and augmentation with $L=550$.

\noindent\textbf{SOLOv2.}
SOLOv2~\cite{wang2020solov2} also resizes the longer image side to $L$. Random cropping is then applied on the resized image with crop size $L \times W = 1333 \times 800$, where $L$ and $W$ always correspond to the longer and shorter side of the resized input image, respectively. Random flipping is applied on top of random cropping as data augmentation. Our \textsc{Disco-Box} with SOLOv2 architecture follows the same data loading and augmentation strategy with SOLOv2.

\begin{figure*}[]
\centering
  \includegraphics[width=\linewidth]{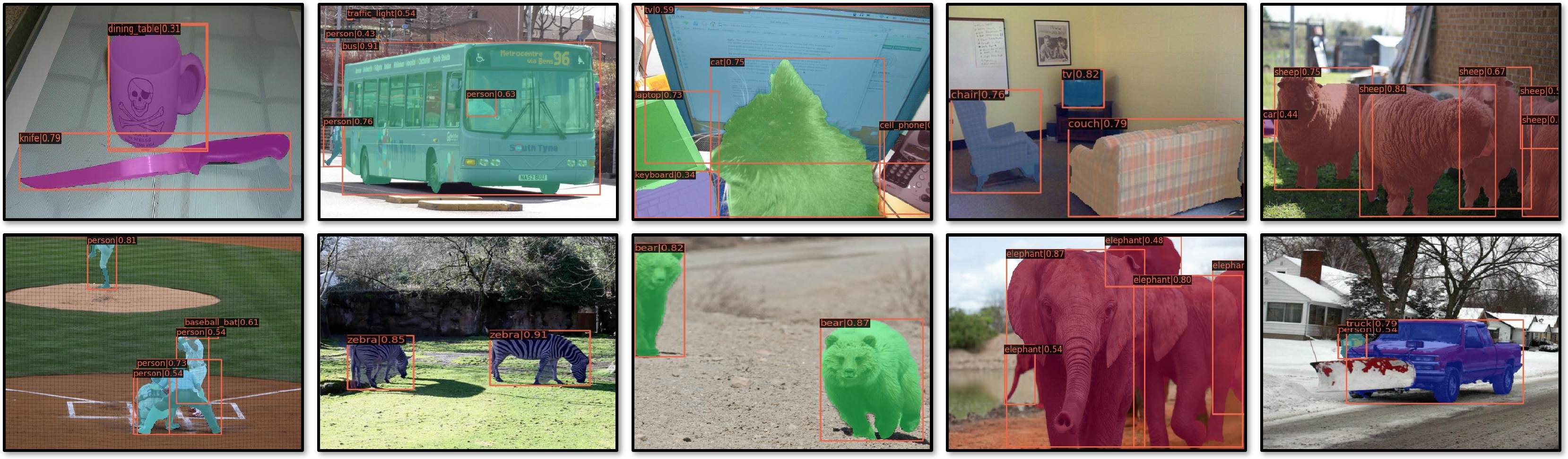}
  \caption{Visualization of instance segmentation on COCO (SOLOv2/ResNeXt-101-DCN).}
\label{fig:vis:coco_inst}
\end{figure*}

\begin{figure*}[]
\centering
  \includegraphics[width=\linewidth]{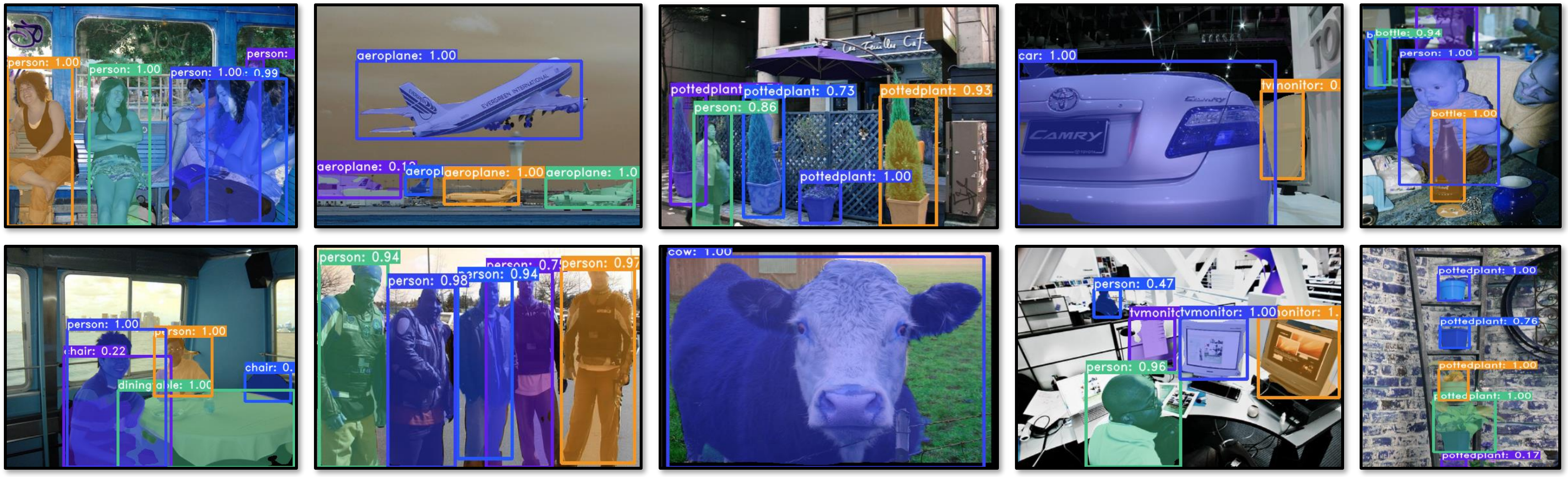}
  \caption{Visualization of instance segmentation on VOC12 (YOLACT++/ResNet-50-DCN).}
\label{fig:vis:voc_inst}
\end{figure*}

\subsection{Learning and optimization}
All our experiments are conducted on DGX-1 machine with 8 NVIDIA Tesla V100 GPUs. Note that we follow the same learning and optimization settings as YOLACT++ and SOLOv2 as elaborated below:

\noindent\textbf{YOLACT++.}
We follow the original setting of~\cite{bolya2020yolact++} for \textsc{Disco-Box} with YOLACT++ architecture. We adopt a batch size of 8 on each GPU. On COCO, the initial learning rate is set to $1\times10^{-3}$ and is decreased to $1\times10^{-4}$ and $1\times10^{-5}$ after 280K and 360K iterations, respectively. Warm-up is used in the first 500 iterations to prevent gradient explosion. On PASCAL VOC 2012, the same warm-up and initial learning rate is applied, and the learning rate is decreased to $1\times10^{-4}$ and $1\times10^{-5}$ after 60K and 100K iterations, respectively. It takes about 5 days to train on COCO and about 12 hours on VOC12.

\noindent\textbf{SOLOv2.}
We follow the original training settings of~\cite{wang2020solov2} where multi-scale training are applied for ResNeXt-101-DCN and ResNet-101-DCN backbones. On COCO, the batch size is set to 2 on each GPU. The initial learning rate is set to $1\times10^{-2}$ and decreased to $1\times10^{-3}$ and $1\times10^{-4}$ after 26 epochs and 32 epochs, respectively. Warm-up is used in the first 2000 iterations. It takes about 4 days to train on COCO with a ResNet-50 backbone.

\subsection{Object retrieval with memory bank}

\noindent\textbf{Pushing objects.}
We maintain an independent first-in-first-out queue with size 100 for every category as the memory bank. The RoI features and predicted masks of objects in each batch are pushed into the memory bank. Note that the RoI features and masks are obtained by the teacher network with respect to the ground truth bounding boxes.

\noindent\textbf{Retrieving objects.} To construct intra-class pairs for every batch during training, we define an object in the current batch as a \textit{query}, and use it to retrieve the RoI features and mask probability maps of intra-class objects from a memory bank. This allows us to conveniently construct pairs without significant extra computation. Note that we empirically set the maximum number of retrieved objects to be 10 for every query object. This is done by random sampling from the memory bank. We ignore the pairs when the bank size (and thus the number of retrieved objects) is smaller than 5.

\section{Additional visualization} \label{sec:add_vis}

Finally, we provide visualization of instance segmentation results which are not presented in the main paper due to limited space. Fig.~\ref{fig:vis:coco_inst} shows the results obtained by \textsc{Disco-Box} SOLOv2/ResNeXt-101-DCN on COCO. In addition, Fig.~\ref{fig:vis:voc_inst} shows the results obtained by YOLACT++/ResNet-50-DCN on VOC12. One could see that the predicted masks are of high qualities in general, even though some examples can be cluttered and challenging.

\end{document}